\documentclass[letterpaper, 10 pt, conference]{ieeeconf}  
\IEEEoverridecommandlockouts                              

\usepackage{graphicx}
\usepackage{amsmath}
\usepackage[flushleft]{threeparttable}
\usepackage{array,booktabs,makecell}
\usepackage[font=small]{caption}
\usepackage{algorithm}
\usepackage{multirow}
\usepackage{algpseudocode}
\usepackage{tikz}
\usepackage{textcomp}
\makeatletter
\newcommand*{\rom}[1]{\expandafter\@slowromancap\romannumeral #1@}
\makeatother

\title{\LARGE \bf
RGB-D Semantic SLAM for Surgical Robot\\
Navigation in the Operating Room}

\author{Cong~Gao$^{1}$, Dinesh~Rabindran$^{2}$, and~Omid~Mohareri$^{2}$
\thanks{This work was supported by Intuitive Surgical}%
\thanks{$^{1}$Johns Hopkins University, Baltimore, United States {\tt\small cgao11@jhu.edu}}
\thanks{$^{2}$Advanced Research, Intuitive Surgical Inc., Sunnyvale, CA, USA}
}

\begin{document}

\maketitle

\begin{abstract}
Gaining spatial awareness of the Operating Room~(OR) for surgical robotic systems is a key technology that can enable intelligent applications aiming at improved OR workflow. In this work, we present a method for semantic dense reconstruction of the OR scene using multiple RGB-D cameras attached and registered to the da Vinci Xi surgical system. We developed a novel SLAM approach for robot pose tracking in dynamic OR environments and dense reconstruction of the static OR table object. We validated our techniques in a mock OR by collecting data sequences with corresponding optical tracking trajectories as ground truth and manually annotated 100 frame segmentation masks. The mean absolute trajectory error is $11.4\pm1.9$~mm and the mean relative pose error is $1.53\pm0.48$ degrees per second. The segmentation DICE score is improved from 0.814 to 0.902 by using our SLAM system compared to single frame. Our approach effectively produces a dense OR table reconstruction in dynamic clinical environments as well as improved semantic segmentation on individual image frames.   

\end{abstract}


\section{Introduction}
Developing artificial intelligence~(AI) systems for surgical operating rooms~(OR) has gained great attention in recent years~\cite{chadebecq2020computer}. Computer vision algorithms have been developed for human pose estimation~\cite{kadkhodamohammadi2017multi, jacob2014context}, surgical activity analysis~\cite{sharghi2020automatic, gholinejad2019surgical}, endoscopic anatomy detection and reconstruction~\cite{ahmad2019artificial, liu2020reconstructing}, and image-based instrument navigation~\cite{fu2021future, gao2019localizing}, etc. With the advancement of robot-assisted minimally-invasive surgeries, the combination of AI and robotics is changing the OR~\cite{andras2020artificial}. Studies have shown that robotic surgeries with AI induced reduction of surgical error rate~\cite{shouval2017machine}, presented higher predictive accuracy~\cite{hashimoto2018artificial}, more precise and reliable interventions~\cite{vercauteren2019cai4cai} and increased efficiency of operating room use~\cite{zhao2019machine}. 

Gaining spatial awareness of the OR environment is crucial for robotic systems to become context aware, provide guidance to the user and potentially automate cumbersome surgical tasks. Several attempts have tried to install cameras inside the OR to track activities~\cite{twinanda2015data, kadkhodamohammadi2017multi, yeung2019computer}. Beyl et al. used 3D cameras mounted on the ceiling of the OR for safe human robot interaction and action perception~\cite{beyl2016time}. Multiple Kinect and Time-of-Flight~(ToF) cameras were rigidly registered to the robot. However, such setup requires the robot arm to be fixed to the OR table for accurate tracking, and thus is not suitable for robots such as the da Vinci Surgical System~(Intuitive Surgical Inc., Sunnyvale, US). Li et al. proposed a 3D OR perception system with multiple ToF cameras installed on the da Vinci patient side cart~(PSC) robot ~\cite{li2020robotic}. Cameras were calibrated to the robot kinematics and semantic segmentation results were fused together. However, the perception ran only on individual frames rather than on video sequences, which did not build the map of the robot motion trajectory and hence limited the quality of 3D point cloud reconstructions.

\begin{figure}[t!]
    \centering
    \includegraphics[width=0.5\textwidth]{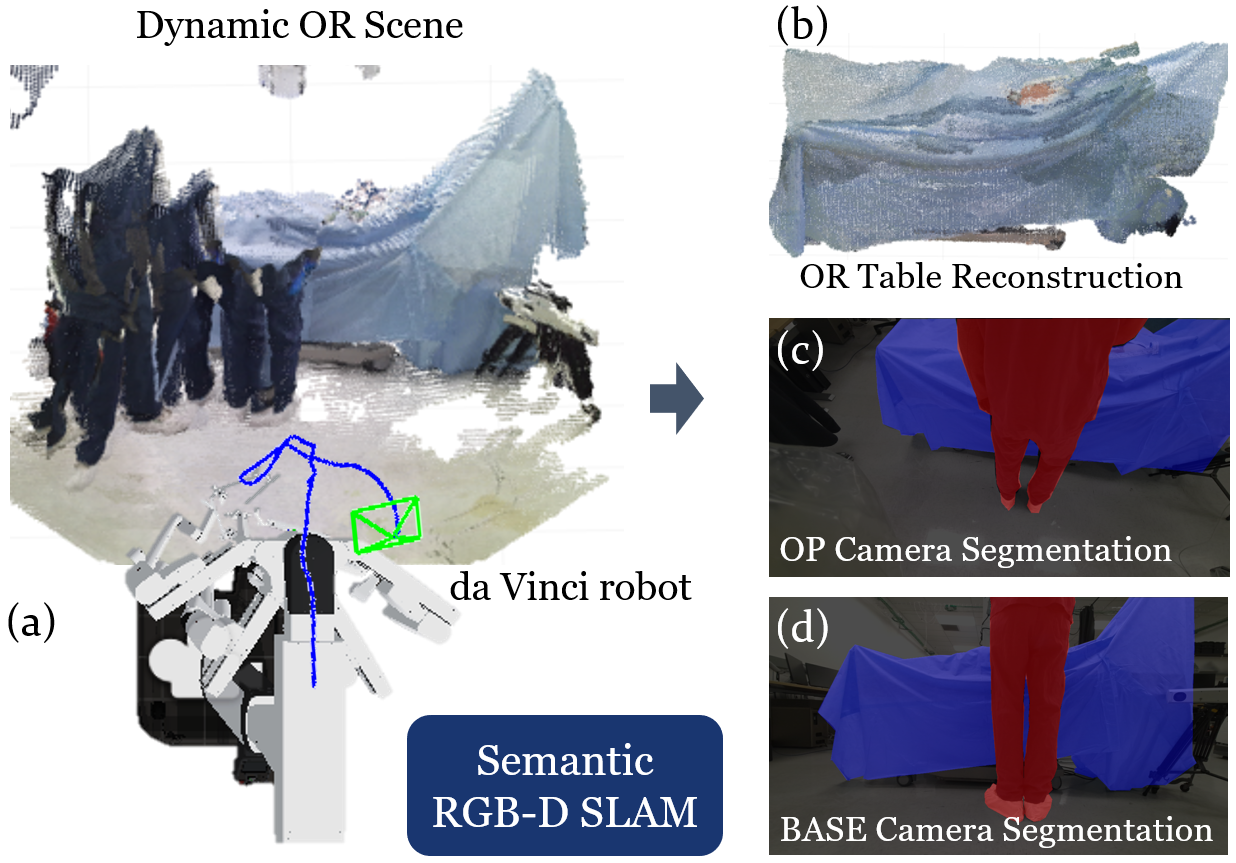}
    \caption{Overall scheme of our semantic RGB-D SLAM. (a) Dynamic OR scene reconstruction using RGB-D SLAM by moving a da Vinci robot. SLAM trajectory estimation is overlayed on the robot model. (b) Semantic OR table reconstruction using our approach. (c)(d) Segmentations of OR table and human overlayed on RGB images from the robot OP and BASE cameras, respectively.}
    \label{fig:capital}
\end{figure}

Visual simultaneous localization and mapping (SLAM) is the technique for camera pose tracking and 3D reconstruction. Given the prior work of Li et al.~\cite{li2020robotic}, we seek to develop a SLAM system to reconstruct dense 3D point clouds from the OR using multiple RGB-D cameras installed on the da Vinci PSC. In this paper, we present a multi-view RGB-D SLAM system to supplement the PSC robot. Two Microsoft Azure Kinect RGB-D cameras are attached to the robot. Our development is based on the state-of-the-art ORB feature based SLAM architecture, ORB-SLAM3~\cite{campos2021orb}. The camera data, including Infrared~(IR) intensity image, depth image and RGB color image, are used for semantic segmentation, tracking and point cloud reconstruction. The cameras are calibrated to the robot kinematics chain via a hand-eye calibration technique. Semantic reconstructions from the two cameras are fused together as a global reconstruction of the OR table. This global reconstruction is then used for improving semantic segmentation on each image frame.

\begin{figure*}[t!]
    \centering
    \includegraphics[width=0.95\textwidth]{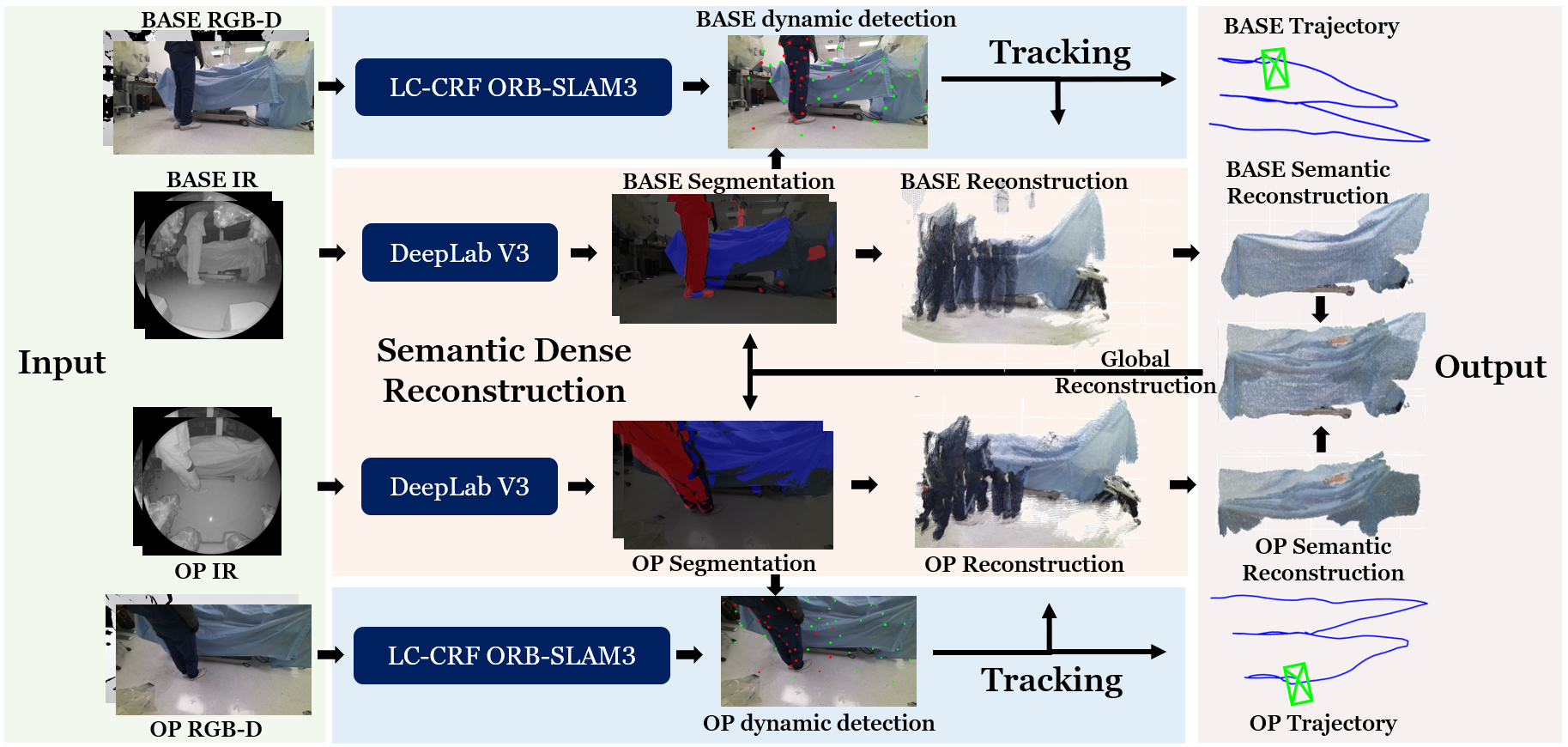}
    \caption{Overall pipeline of our system. Input: IR and RGB-D image streams from BASE and OP cameras. Tracking: The RGB-D images are used for LC-CRF ORB SLAM tracking, which generates camera trajectory and point cloud reconstruction. Semantic Dense Reconstruction: IR images pass through DeepLab V3~\cite{chen2018encoder}, which produces semantic segmentation. The segmentation helps with dynamic landmark detection for tracking and filters the point cloud to become semantic reconstruction. Output: camera trajectories and fused global reconstruction. The global semantic reconstruction can further improve the image segmentation.}
    \label{fig:pipeline}
\end{figure*}

OR is a complex indoor environment with high risk. The scene includes both dynamic objects~(like the OR staff) and static objects~(like the OR table). The OR lighting conditions vary a lot between pre-operative~(bright) and intra-operative operations~(dark) phases. The unique scenario of OR makes it challenging for accurate tracking and clean reconstruction. The OR table is of special interest since it is a target object for robot setup and docking before the surgical procedure.  Du et al. proposed a dynamic SLAM method based on ORB-SLAM2 using CRF-based long-term consistency~(LC-CRF)~\cite{du2020accurate}, which detects the ORB feature points as dynamic or static. It shows superior performance on the environments including dynamic object motions. We combined the LC-CRF SLAM dynamic point detections with our improved semantic segmentation to achieve accurate SLAM trajectory tracking in the OR.

Our contributions in this work are summarized as follows: We developed an RGB-D semantic SLAM system for the da Vinci robot navigation in the OR. Our system used multiple RGB-D cameras calibrated to the robotic kinematics chain. We integrated the approach of ORB-SLAM3 with CRF-based long-term consistency to densely reconstruct the static OR table object, improve image segmentation, and estimate the robot trajectory. We manually collected data sequences in a mock OR environment with optical tracker trajectories and annotated segmentation labels as ground truth for our evaluations. 

\section{Methodology}
Our system pipeline is shown in Fig.~\ref{fig:pipeline}. The system uses two Azure Kinect RGB-D sensors~(Microsoft, USA) calibrated to the kinematics chain of a da Vinci Xi PSC robot. It takes the camera's infrared~(IR) images, depth images and RGB color images as input, simultaneously runs two SLAM modules for real-time tracking, and performs learning-based semantic segmentation, semantic point cloud reconstruction and fusion. The IR images are passed through a pre-trained DeepLab V3~\cite{chen2018encoder} segmentation model to generate segmentation masks. Both RGB and depth images are used for RGB-D SLAM tracking. Dense point cloud reconstructed from the depth image is filtered by the segmentation. A global point cloud is maintained by fusing the two semantic point clouds. The segmentations are further refined by projecting the global point cloud to each camera frame. Details of the pipeline are described in the following sections.

\subsection{System Setup and Calibration}
\begin{figure}[t!]
    \centering
    \includegraphics[width=0.45\textwidth]{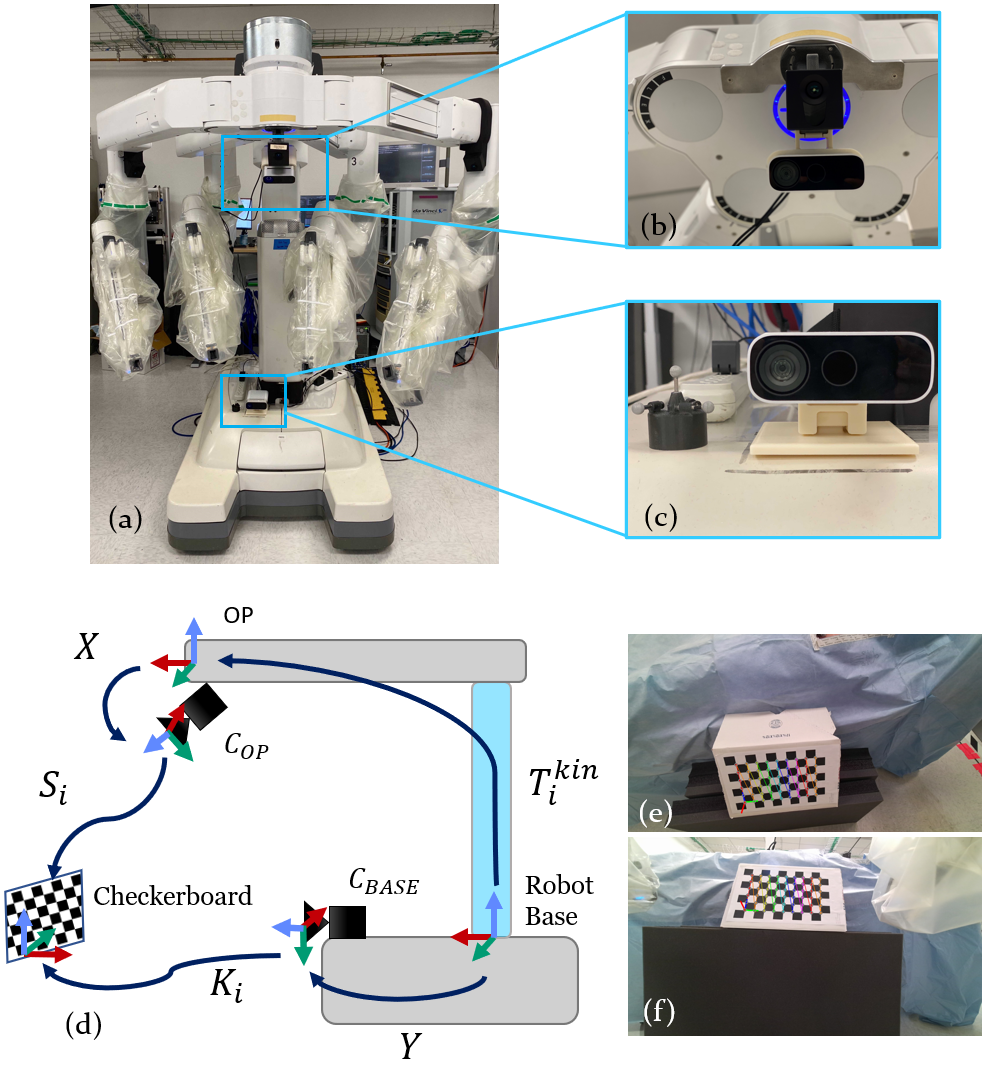}
    \caption{Robotic system setup and illustration of the hand-eye calibration. (a) Our da Vinci robotic system with Kinect sensors. (b) OP Kinect camera. (c) BASE Kinect camera and optical marker. (d) Hand-eye calibration transformations. The dark blue arrows are rigid transformations. The cross arrows in RGB colors represent the coordinate frames. (e) An example calibration image from the OP Kinect camera. (f) The calibration image from the BASE Kinect camera, which corresponds to the image in (e).}
    \label{fig:calibration}
\end{figure}

The Kinect sensor has two cameras to produce RGB-D image pairs, including a Time-of-Flight~(ToF) depth camera and an RGB camera. IR and depth images generated from the ToF camera are transformed to the RGB camera frame (resolution $720\times1280$) using the sensor's factory calibration parameters. Dense 3D point cloud is reconstructed from the depth image using the camera intrinsics. Two Kinect sensors are mounted on the da Vinci Xi PSC robot Orienting Platform~(OP) and BASE locations, respectively. We denote these two Kinect sensors as $C_{OP}$ and $C_{BASE}$. The 3D printed mount parts are custom-designed to ensure the rigid attachment of the Kinect cameras to the robot joints~(Fig.~\ref{fig:calibration}(a)-(c)).

We performed hand-eye calibration to register the camera data to the robot kinematics chain. As shown in Fig.~\ref{fig:calibration}(d), a checkerboard of size $6\times9$ was printed and fixed in front of the robot. The unknown calibration matrices, $X$ and $Y$, refer to transformations from OP to $C_{OP}$ and Robot Base to $C_{BASE}$, respectively. The robot base was static during the calibration procedure. We moved the OP camera~($C_{OP}$) to various static configurations within the robot work space by changing the robot kinematics. At configuration $i$, we recorded the robot forward kinematics~($T_i^{kin}$) and the RGB image. The pose of $C_{OP}$ relative to the checkerboard~($S_i$) was estimated by first detecting the checkerboard pattern and then solving the PnP problem. Given two arbitrary configurations $i$, $j$, we have
\begin{align}
    S_i X T_i^{kin} & = S_j X T_j^{kin}, \\
    S_j^{-1} S_i X & = X T_j^{kin} (T_i^{kin})^{-1}.
\end{align}
This is essentially an $AX=XB$ problem, where $A=S_j^{-1} S_i$ and $B=T_j^{kin} (T_i^{kin})^{-1}$. We used the method of~\cite{andreff1999line} to solve $X$. Because $C_{BASE}$ is fixed relative to robot base, $Y$ can not be solved using the same routine. However, with the knowledge of $X$, we positioned the checkerboard to be within the capture range of both $C_{OP}$ and $C_{BASE}$~(Fig.~\ref{fig:calibration}(e)(f)). $K_i$ is the pose of $C_{BASE}$ relative to the checkerboard. Then, $Y$ is solved using:
\begin{equation}
    Y = K_i^{-1} S_i X T_i^{kin}.
\end{equation}
To this end, we have calibrated $C_{OP}$ and $C_{BASE}$ to the robot kinematics chain, which means data retrieved from the cameras can all be transformed to the robot base frame. The calibration is used to fuse the point cloud reconstructions in Section~\rom{3} D.

\subsection{Dynamic RGB-D SLAM}
Our RGB-D SLAM tracking system is developed based on the ORB feature-based SLAM architecture, ORB-SLAM3~\cite{campos2021orb}. For each incoming frame~($f_c$), the RGB image~($I^{rgb}_c$) and depth image~($I^{depth}_c$) serve as input RGB-D data for tracking the camera pose~($T_{wc}$), where $w$ denotes the world frame and $c$ the current frame corresponding to the associate camera. Key frames~($f_k$) are saved in the SLAM optimization model, and are used to optimize the map via bundle adjustments upon loop closure detection. 

Because the OR scene lighting conditions vary between pre-operative and intra-operative operations, we applied a semantic segmentation model~($M^{seg}$), which is a variation of the DeepLab V3+ developed by Li et al.~\cite{li2020robotic}, for OR scene segmentation. This model takes the IR image~($I^{IR}_c$) as input and thus is invariant to the natural lighting conditions. The model predicts a label mask~($I^{mask}_c$) for multiple OR objects including OR table, human, ceiling light, mayo stand, sterile table, PSC, etc. In this work, we take the OR table as the representative static object of interest and the human as moving object to demonstrate the SLAM system. 

The conventional ORB-SLAM3 uses frame-by-frame corresponding ORB feature landmarks to estimate the camera trajectory. However, it lacks the ability to differentiate static/dynamic ORB features, which makes the tracking result less accurate in highly dynamic scene. We integrated the method of dynamic SLAM using CRF-based long-term consistency~(LC-CRF)~\cite{du2020accurate} to our tracking system. It uses a CRF with appropriate unary and pairwise potentials to label each landmark as static or dynamic. Labeled dynamic landmarks are rejected from the SLAM computations. With the knowledge of semantic segmentation mask, we further classified the ORB feature landmarks attributed to known moving objects like human as dynamic ORB features.

Our dynamic RGB-D SLAM runs concurrently on both the BASE and OP Kinect cameras, which we denote as BASE SLAM and OP SLAM, respectively. Because the BASE camera is rigidly mounted on the robot base, we used the BASE SLAM tracking for robot trajectory estimation. The tracking results from both BASE and OP SLAMs are used for fusing semantic point cloud reconstructions.

\subsection{Semantic Reconstruction}
\begin{algorithm}[h!]
\caption{Semantic Reconstruction}\label{alg:sem-recon}
\textbf{Input:}\\
\hspace*{\algorithmicindent}Kinect camera data: RGB image $I^k_{rgb}$, depth image $I^{depth}_k$, segmentation mask $I^k_{mask}$; camera intrinsic parameters: $\theta_{cam}$.\\
\textbf{Output:}\\ 
\hspace*{\algorithmicindent}Semantic dense point cloud $P_k$.
\begin{algorithmic}[1]
\item Get key frame camera pose $T_{wk} = Tracking(I^k_{rgb}, I^{depth}_k)$.
\item $I^{mask}_k = (I^{mask}_k \ominus b) \oplus b$
\item $I^{depth}_k = Filter(I^{depth}_k, I^{mask}_k)$
\item $P_k = Reconstruct(I^{depth}_k, \theta_{cam})$
\item $P_k = T_{wk}\cdot P_k$ \Comment{Transform $P_k$ to world frame}
\If{$k$ is 1}
    \State $P^{cam} = P_k$
\ElsIf{$k>1$}
    \State $P_k = ICP(P^{cam}, P_k)$
    \State $P^{cam} = P^{cam} + P_k$
    \State $P^{cam} = OutlierReject(P^{cam})$
    \State $P^{cam} = VoxelResampling(P^{cam})$
\EndIf
\end{algorithmic}
\end{algorithm}
Our SLAM system aims at building a dense semantic reconstruction of the OR table. A dense point cloud~($P_k$) was reconstructed from the depth image~($I^{depth}_k$) using the camera manufacturing intrinsic parameters~($\theta_{cam}$) at each key frame $f_k$. The IR image~($I^{IR}_k$) was downsampled from its natural resolution of $512\times 512$ to $256\times 256$ and passed through the segmentation model to generate a prediction mask image: $I^{mask}_k = M^{seg}(I^{IR}_k)$. To refine the mask, we performed an opening morphological operation to reject outlier predictions: $(I^{mask}_k \ominus b) \oplus b$, where $b$ is a circular structure with a radius of 10 pixels. $I^{mask}_k$ was then used to filter $P_k$ based on pixel-wise OR table predictions. $P_k$ was transformed to the SLAM world coordinate frame using the camera pose estimation $T_{wk}$ from the SLAM tracking module. A camera global point cloud~($P^{cam}, cam\in\{OP, BASE\}$) in the SLAM world coordinate frame was maintained by fusing incoming $P_k$. We performed an ICP registration to fuse $P_k$ to $P^{cam}$ followed by a statistical outlier rejection. The ICP registration was designed to be a local adjustment, so we set a maximum of 10 iteration steps for speed concerns. To remove outlier point clusters, we calculated mean k-nearest neighbor distances~(k = 100) for every point neighborhood, and rejected all points which have a distance larger than 3 standard deviations of the mean distance to the query point. Finally, $P^{cam}$ was downsampled based on 3D voxel grid to keep its uniform density, where the voxel resolution was set to 10 mm isotropic. This procedure is summarized in Algorithm~\ref{alg:sem-recon}.  

The semantic reconstruction $P^{cam}$ accumulates the knowledge of the OR table object as the robot is being moved, which is used to improve the incoming individual frame segmentation performance. We performed a reprojection of the $P^{cam}$ to the camera image plane once it was transformed from the world frame to the current camera frame: $I^{mask, cam}_c = Reproject(T_{wc}^{-1}\cdot P^{cam}_c)$. This refined segmentation is considered better than the single frame network segmentation for two reasons: 1) $P^{cam}$ keeps the accumulated historic object information; 2) the outlier rejection of $P^{cam}$ keeps the object shape and removes noisy points in 3D space. However, $P^{cam}$ is limited by the single camera capture range. A more complete reconstruction is achieved by fusing multiple SLAMs. 

\subsection{Multi-SLAM Fusion}
Semantic reconstructions from BASE and OP SLAM modules were fused together to create a global reconstruction~($P^{global}$). Given the hand-eye calibration matrices $X$ and $Y$, $P^{cam}$ in each SLAM world frame was transformed to the robot base frame following the chain of transformation in Fig.~\ref{fig:calibration}(d) by using the current forward robot kinematics~($T^{kin}_c$). We performed an ICP registration between these point clouds, followed by an outlier rejection and voxel grid downsampling described in Section~\rom{3} C. $P^{global}$ was reprojected to each individual camera image planes to refine the segmentation. The multi-SLAM fusion process is described in Algorithm~\ref{alg:multi-fuse}.

The fused global reconstruction merges information from different perspectives, which compensates each individual camera's limited field of view. The reprojected segmentation is more coherent and robust to outliers. Performing multi-SLAM fusion also improves the efficiency of gaining spatial awareness, which intuitively reconstructs the object in only a few steps. 

\begin{algorithm}[t!]
\caption{Multi-SLAM Fusion}\label{alg:multi-fuse}
\textbf{Input:}\\
\hspace*{\algorithmicindent}OP camera segmentation and reconstruction: $I^{mask, OP}_c$, $P^{OP}$, BASE camera reconstruction: $I^{mask, BASE}_c, P^{BASE}$, Robot forward kinematics: $T_c^{kin}$, Calibration matrices: $X, Y$.\\
\textbf{Output:}\\ 
\hspace*{\algorithmicindent}Semantic global dense point cloud $P^{global}$.
\begin{algorithmic}[1]
\item $P^{OP} = (T_c^{kin})^{-1}\cdot X^{-1}\cdot P^{OP}$\Comment{Transform $P^{OP}$ to robot base frame}
\item $P^{BASE} = Y^{-1}\cdot P^{BASE}$\Comment{Transform $P^{BASE}$ to robot base frame}
\item $P^{OP} = ICP(P^{BASE}, P^{OP})$
\item $P^{global} = P^{BASE} + P^{OP}$
\item $P^{global} = OutlierReject(P^{global})$
\item $P^{global} = VoxelResampling(P^{global})$
\item $I^{mask, OP}_c = Reproject(Y\cdot P^{global})$
\item $I^{mask, BASE}_c = Reproject(X\cdot T_c^{kin}\cdot P^{global})$
\end{algorithmic}
\end{algorithm}

\section{Experiments and Results}
We validated our approach using a manually collected dataset with ground truth. We considered the aspects of SLAM trajectory estimation accuracy, OR table segmentation accuracy and qualitative results of point cloud reconstruction. In this section, we describe the data collection process and each evaluation metric with discussions. Of note, because our study is proof-of-concept, we did not optimize the implementation to supply real-time application.

\begin{figure*}[t!]
    \centering
    \includegraphics[width=\textwidth]{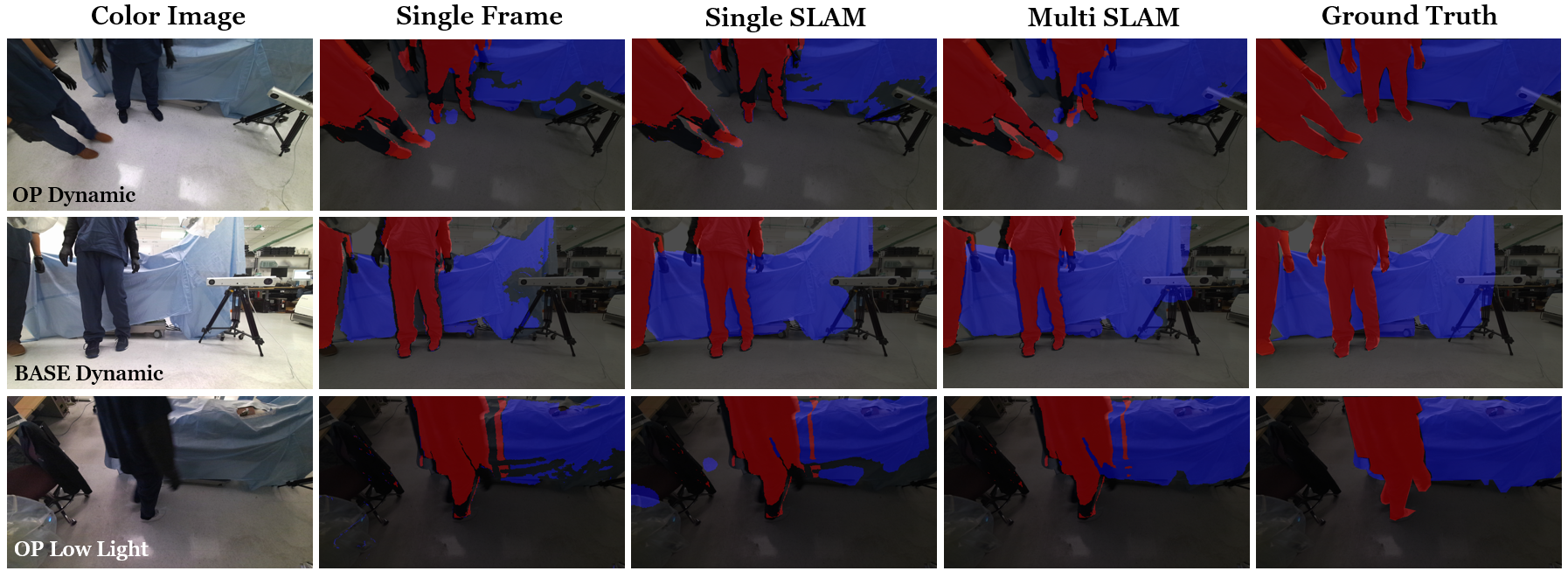}
    \caption{Qualitative segmentation results. Three representative frames are selected from OP camera dynamic scene, BASE camera dynamic scene and OP camera low light scene. Each row presents the corresponding color image, blended segmentation overlays of single frame network segmentation, single camera SLAM reconstruction reprojected segmentation, fused SLAM global reconstruction reprojected segmentation and manually annotated ground truth. The OR table segmentation is blended in blue and the human segmentation in red.}
    \label{fig:segmentation}
\end{figure*}

\subsection{Data Collection}
We collected time-synchronized sensor (RGB, IR and depth images) and robot kinematics data in a clinical development lab. The environment was setup similar to real clinical settings by placing a draped OR table in the center of the lab. The robot was manually moved by a user to simulate the da Vinci setup and docking procedure and multiple sequences have been captured under various conditions, including normal light static scene, normal light dynamic scene, and mixed static/dynamic scene under low lighting conditions. In dynamic scene, we simulated human motion by having multiple people walking around in the room while we moved the robot. The low lighting condition was to mimic the intra-operative OR scenario. In order to validate the accuracy of the camera trajectory estimation, a polaris fiducial marker was fixed to the base of the robot~(Fig.~\ref{fig:calibration}(c)). We recorded the marker's trajectories using an Atracsys Optical Tracking System~(Atracsys LLC, Switzerland), which were used as the ground truth for validation. Captured images, robotic kinematics and optical tracking data were time synchronized. Image frames selected from the sequences were manually labeled to evaluate the segmentation performance.

\subsection{Robot Trajectory}
\begin{table}[b!]
  \caption{Robot Trajectory Accuracy}
  \label{table:trajectory}
  \centering 
  \begin{threeparttable}
    \begin{tabular}{ccccc}
         & & Static & Dynamic & Low light\\
     \midrule\midrule
     \multirow{2}{*}{\shortstack{STD ATE\\(mm)}} & Mean & $11.6\pm 2.7$ & $11.5\pm2.3$  & $10.9\pm1.6$\\
                                  & Median & 8.6 & 8.2 & 9.6 \\
     \hline
     \multirow{2}{*}{\shortstack{CRF ATE\\(mm)}} & Mean & $11.5\pm2.4$ & $11.4\pm1.9$ & $11.2\pm2.1$ \\
                                   & Median & 8.7 & 7.9 & 10.1\\
     \hline
     \multirow{2}{*}{\shortstack{STD RPE\\($^\circ$)}} & Mean & $1.41\pm 0.74$ & $1.24\pm0.14$  & $1.76\pm0.65$ \\
                                   & Median & 0.65 & 0.64 & 0.48 \\
     \hline
     \multirow{2}{*}{\shortstack{CRF RPE\\($^\circ$)}} & Mean & $1.62\pm0.49$ & $1.25\pm1.13$ & $1.53\pm0.77$ \\
                                  & Median & 0.65 & 0.65 & 0.47\\
     \bottomrule
    \end{tabular}
  \end{threeparttable}
  \begin{tablenotes}
     \item Note: STD refers to standard ORBSLAM3. CRF refers to LC-CRF ORBSLAM3.
   \end{tablenotes}
\end{table}

The trajectory evaluation uses two metrics to measure the accuracy between the robot trajectory estimation and the ground truth: the absolute trajectory error~(ATE, measured in millimeters) and the relative pose error~(RPE, measured as rotation in degrees per second), as defined in~\cite{sturm2012benchmark}. The robot trajectories were estimated using the BASE camera SLAM, and the ground truth were the optical marker trajectories. We evaluated 15 manually collected trajectories in three scene conditions~(5 trajectories each): static~(normal light), dynamic~(normal light) and low light. The numeric results are summarized in Table.~\ref{table:trajectory}. Our SLAM system achieves a mean of $11.4\pm1.9$~mm ATE and $1.53\pm0.48^\circ$ RPE in all testing trajectories. The result is consistent over static, dynamic and low light conditions, with mean ATE of 11.6~mm, 11.5~mm and 10.9~mm, respectively, and the RPE less than 2 degrees. This accuracy is promising for robot pose estimation and navigation in the OR scene. However, the performance of LC-CRF ORB-SLAM is similar to the standard ORB-SLAM. This is likely because the range of robot motion is limited and the dynamic human motion does not occupy too much of the scene in our collected dataset. In the future, we will consider testing with larger robot movements and more complex scene.  

\subsection{OR Table Segmentation}
\begin{table}[b!]
  \caption{OR Table Segmentation DICE Score}
  \label{table:segmentation}
  \centering 
  \begin{threeparttable}
    \begin{tabular}{ccc}
      & Mean & Median\\
     \midrule\midrule
     BASE Single Frame & $0.827\pm0.072$ & 0.826\\
     OP Single Frame & $0.801\pm0.114$ & 0.836\\
     All Single Frame & $0.814\pm0.096$ & 0.828\\ 
     \hline
     BASE SLAM & $0.886\pm0.062$ & 0.918\\
     OP SLAM & $0.830\pm0.085$ & 0.843\\
     All Cam SLAM & $0.858\pm0.080$ & 0.861\\
     \hline
     All Fused SLAM & $\mathbf{0.902\pm0.054}$ & $\mathbf{0.914}$\\
     \bottomrule
    \end{tabular}
\end{threeparttable}
\end{table}

\begin{figure*}[t!]
    \centering
    \includegraphics[width=\textwidth]{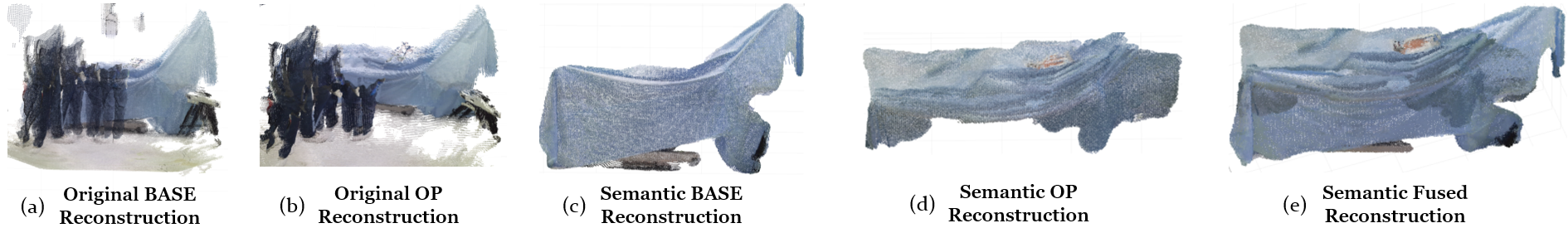}
    \caption{Qualitative comparison of point cloud reconstructions. (a)(b) Examples of original SLAM reconstruction without semantics using BASE and OP SLAMs. (c)(d) Examples of semantic reconstruction using BASE and OP SLAMs. (e) Example of fused semantic reconstruction.}
    \label{fig:quali-pointcloud}
\end{figure*}

We evaluated the segmentation improvement by reporting DICE score of the OR table segmentation on randomly selected frames in our collected dataset. In total, 100 frames consisting of 50 BASE and 50 OP camera images were randomly selected from three trajectory sequences, covering static, dynamic and low light scenes. The OR table label mask was manually annotated on each frame as ground truth. We compared the performance of single frame network segmentation, reprojected segmentation from single camera SLAM reconstruction and reprojected segmentation from fused SLAM global reconstruction. Qualitative results are presented in Fig.~\ref{fig:segmentation}, including examples from BASE and OP cameras in dynamic scene and low light conditions. The segmentations are visually more coherent from reprojected SLAM reconstructions as shown in the figure.

Quantitative results are summarized in Table.~\ref{table:segmentation}. We reported mean and median DICE scores separately on BASE and OP frames for single frame and single camera SLAM methods, and also computed the metric on all the frames together. Performance of BASE Single Frame is derived from the pre-trained model in~\cite{li2020robotic} as baseline. Segmentation using fused SLAM reprojection achieves the highest accuracy of $0.902\pm0.054$. We can see consecutive improvements from a mean DICE score of 0.814, 0.858 to 0.902 and decreasing standard deviations of 0.096, 0.080 to 0.054 by single frame, single camera SLAM and fused SLAM, respectively. This suggests that the semantic SLAM reconstruction makes the segmentation more stable, and the fusion of multiple views further improves the performance. We also notice that the SLAM improvements of the OP camera data is more significant than the BASE camera. This is because the viewing perspective of the OP camera changes more drastically as the robot moves compared to the BASE camera, which makes the OP segmentation benefit more from the 3D reconstruction.   

\subsection{Point Cloud Quality}

\begin{figure}[b!]
    \centering
    \includegraphics[width=0.40\textwidth]{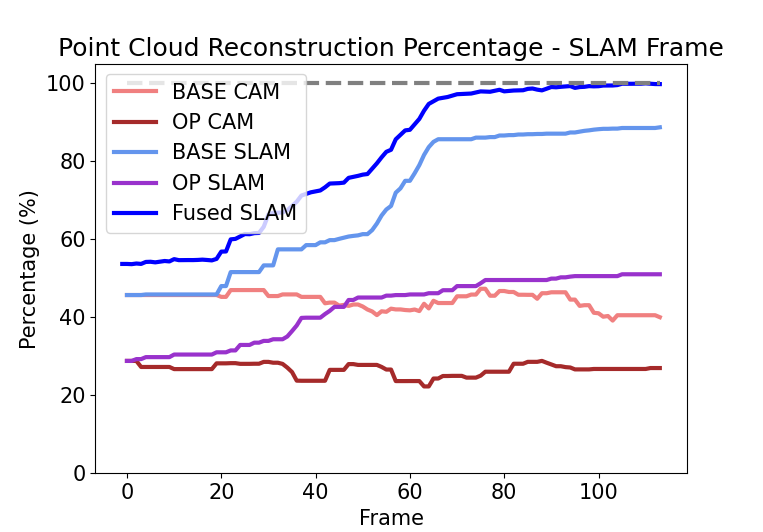}
    \caption{Illustration of increased OR table object reconstruction percentage using SLAM. The dotted gray line represents a complete object reconstruction as 100 percent. The colored curves show the change of object reconstruction percentage compared to the complete one with respect to the SLAM data frame.}
    \label{fig:recon-percentage}
\end{figure}
A qualitative example comparison of the original point cloud and the fused dense point cloud reconstruction is presented in Fig.~\ref{fig:quali-pointcloud}. The original point cloud reconstructed using single RGBD camera by ORB-SLAM includes noisy background and moving objects, such as the human, and its filed of view is limited by the capture range of the single camera. It is thus hard to extract meaningful and stable information. The fused point cloud reconstructed using our method is dense, clean, and has a complete shape of the table. Spatial information of the static object can then be extracted and used to navigate the surgical robot. 

By fusing the point cloud from multiple cameras, the reconstruction efficiency is improved compared to using single camera. We demonstrate this characteristic using the accumulated curvature of the point cloud as a percentage ratio to a complete reconstruction with respect to the SLAM data frame~(Fig.~\ref{fig:recon-percentage}). We used the full point cloud reconstruction size from the fused SLAM as benchmark. The percentage is calculated by dividing the current point cloud size of each algorithm to the benchmark size. We compared the performance of using single camera frame, single camera SLAM and fused SLAM. Single camera frame does not make use of the knowledge in history motion, so the curve is flattened. Due to the capture range limitation, SLAM reconstruction from each individual camera is always less complete than the fused one. This efficiency of object reconstruction is important in the OR because the robot setup can be timely sensitive in emergency cases. 


\section{Conclusion}
In this work, we present an RGB-D semantic SLAM system for a da Vinci robotic OR. The system is used for robot trajectory tracking in a dynamic OR environment and generates dense semantic reconstruction of the OR table object of interest. The pipeline was validated using data collected from a mock OR lab with optical tracking and manual labeling as ground truth. The results suggest that our SLAM system largely improves the robotic perception of the OR environment with accurate trajectory estimation and semantic scene understanding. The system has the potential to assist robot setup, surgical navigation and improve human robot interaction.

\section*{Acknowledgment}
The authors would like to thank Tobias Czempiel, Ali Mottaghi and Zhuohong He for their help during the dataset collection.

\bibliographystyle{IEEEtran}
\bibliography{IEEEabrv,reference.bib}

\end{document}